\documentclass[letterpaper, 10 pt, conference]{ieeeconf}  

\IEEEoverridecommandlockouts                              

\usepackage{xcolor}
\usepackage{gensymb}
\usepackage{graphicx} 
\usepackage{amssymb}

\title{\LARGE \bf
An Image-Guided Robotic System for Transcranial Magnetic Stimulation: System Development and Experimental Evaluation
}

\author{
Yihao Liu $^{1,2,*}$, Jiaming Zhang$^{1,2}$, Letian Ai $^{1}$, \\
Jing Tian $^{2}$, Shahriar Sefati $^{3}$, Huan Liu $^{4}$, Alejandro Martin-Gomez $^{1,6}$, \\ 
Amir Kheradmand $^{2,7,**}$ and Mehran Armand $^{1,3,5,**}$
\thanks{Manuscript submitted Sept. 15, 2024. This work was supported by a grant from the National Institute of Deafness and Other Communication Disorders (R01DC018815) and internal funding from the Johns Hopkins University Department of Computer Science. }
\thanks{$^{1}$ Dept. of Computer Science, JHU, Baltimore, MD, USA}%
\thanks{$^{2}$ Dept. of Neurology, JH Medicine, Baltimore, MD, USA}
\thanks{$^{3}$ Dept. of Mechanical Eng., JHU, Baltimore, MD, USA}
\thanks{$^{4}$ Sch. of Autom., China U. of Geosci., Wuhan, Hubei, China}
\thanks{$^{5}$ I3R, U of Arkansas, Fayetteville, AK, USA}
\thanks{$^{6}$ Dept. of EECS, U of Arkansas, Fayetteville, AK, USA}
\thanks{$^{7}$ Dept. of Neuroscience, JH Medicine, Baltimore, MD, USA}
\thanks{$^{*}$ Corresponding author {\tt\small (yliu333@jhu.edu)}}
\thanks{$^{**}$ Equal contribution as senior authors}
}

\begin{document}

\maketitle
\thispagestyle{empty}
\pagestyle{empty}

\begin{abstract}

Transcranial magnetic stimulation (TMS) is a noninvasive medical procedure that can modulate brain activity, and it is widely used in neuroscience and neurology research. Compared to manual operators, robots may improve the outcome of TMS due to their superior accuracy and repeatability. However, there has not been a widely accepted standard protocol for performing robotic TMS using fine-segmented brain images, resulting in arbitrary planned angles with respect to the true boundaries of the modulated cortex. Given that the recent study in TMS simulation suggests a noticeable difference in outcomes when using different anatomical details, cortical shape should play a more significant role in deciding the optimal TMS coil pose. In this work, we introduce an image-guided robotic system for TMS that focuses on (1) establishing standardized planning methods and heuristics to define a reference (true zero) for the coil poses and (2) solving the issue that the manual coil placement requires expert hand-eye coordination which often leading to low repeatability of the experiments. To validate the design of our robotic system, a phantom study and a preliminary human subject study were performed. Our results show that the robotic method can half the positional error and improve the rotational accuracy by up to two orders of magnitude. The accuracy is proven to be repeatable because the standard deviation of multiple trials is lowered by an order of magnitude. The improved actuation accuracy successfully translates to the TMS application, with a higher and more stable induced voltage in magnetic field sensors. 

\end{abstract}


\begin{figure*}[ht]
    \centering
    \includegraphics[width=\textwidth]{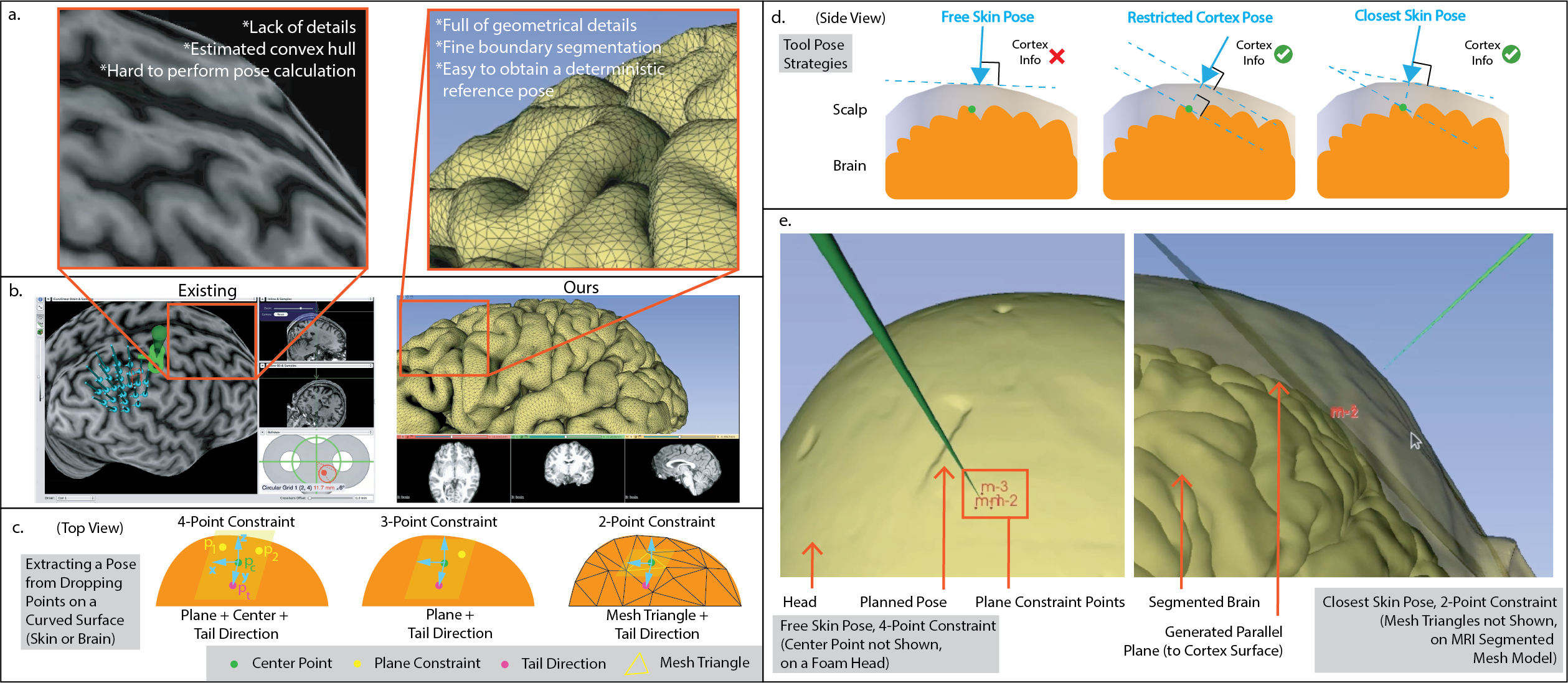}
    \caption{Neuro-navigation system and pre-operative planning strategies. \textbf{Panels a. and b.} illustrate the comparison between our proposed system and an existing commercial system. Our system uses a fine-segmented brain mesh model to maintain geometrical details. This information can be used to determine the approaching pose of the TMS coil so a reference (true zero) can be established. In contrast, the existing commercial system uses a radially sliced volume image to estimate the boundary of the brain. It lacks details, and the pose of the TMS coil cannot be automatically calculated based on the true geometry. \textbf{Panel c.} shows the strategies to automatically calculate a pose on a mesh surface (Section \ref{sec:toolplansurface}). Each of these strategies defines both a position, determined by a center point and an orientation, which is established in different ways. The \textit{4-point constraint} uses a center point along with three additional points, including one that determines the tail direction, to form a plane that defines the orientation. The \textit{3-point constraint} selects one of three points as the center position. Lastly, the \textit{2-point constraint} searches the triangle within the mesh model that contains the specified center point to determine the orientation. Using any of the above constraints, \textbf{Panel d.} shows three different TMS coil pose calculation methods (Section \ref{sec:toolplanstrategies}). \textit{Free Skin Pose} uses the shape of the skin and gets a pose perpendicular to the scalp. It uses no cortical geometrical information except the human estimation of the target's location. \textit{Restricted Cortex Pose} obtains an orientation tangential to the target cortex. \textit{Closest Skin Pose} uses the target location in the cortex, projects it to the closest point on the skin, and obtains a tangential pose to the skin. Note these three strategies are \textit{not} individually corresponding to the strategies in panel c. Each strategy in panel d. may use any pose calculation method in panel c. \textbf{Panel e.} shows examples using our proposed system.}
    \label{fig:preopplan}
\end{figure*}

\begin{figure*}[ht]
    \centering
    \includegraphics[width=\textwidth]{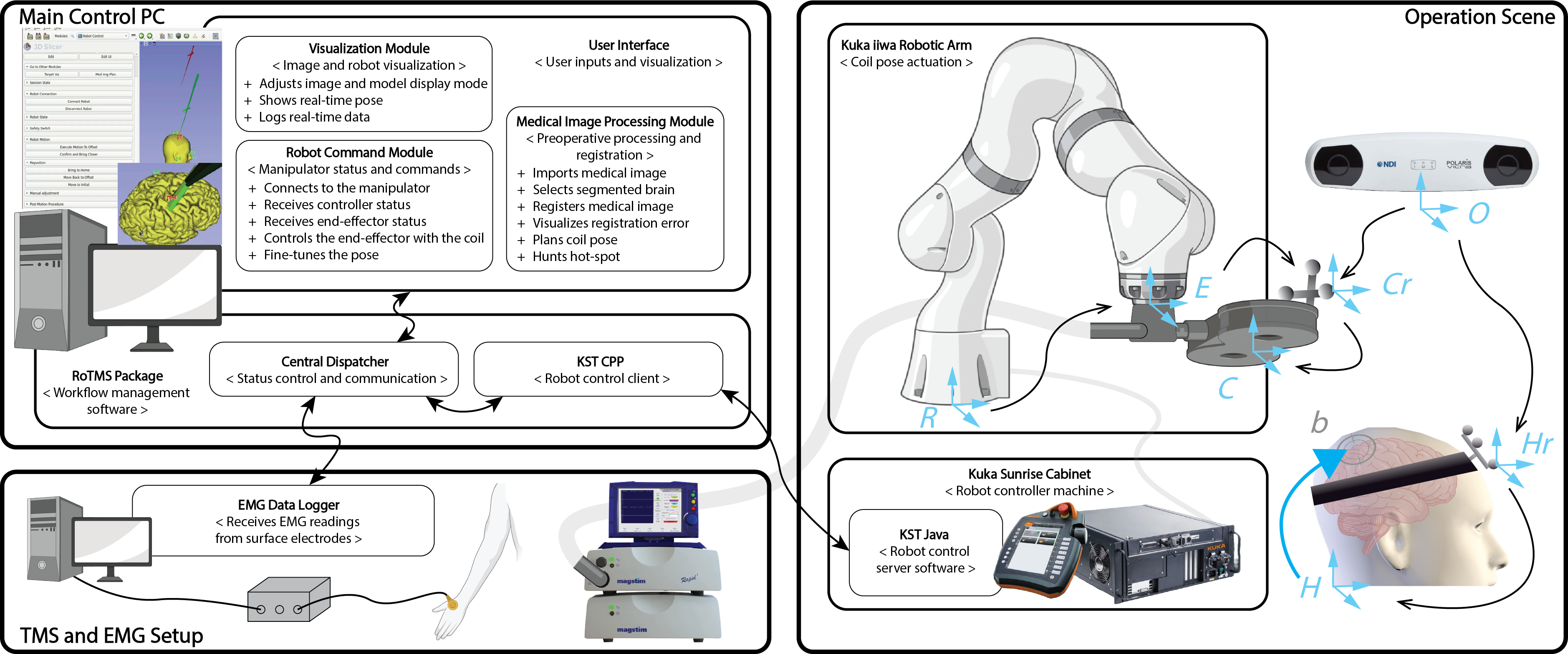}
    \caption{The architecture and the kinematics of the proposed image-guided robotic TMS system. The \textbf{Main Control PC} hosts the user interface and the robotic TMS (RoTMS) control packages. The user interface visualizes the medical images, segmented models, and real-time poses of the robot end-effector. The pre-operative plans and the robot command are also entered from the user interface. The RoTMS package manages the workflow and communications between the controller and the user interface. The \textbf{Operation Scene} contains a robotic arm, a controller, an optical tracker, and a human subject with a rigid body marker attached to the head. Arrows and reference frames illustrate the complete kinematic chain. From left to right, these reference frames are \{\textit{R}\} robot base, \{\textit{E}\} end-effector, \{\textit{C}\} center of the TMS coil, \{\textit{Cr}\} rigid body marker on the TMS coil, \{\textit{O}\} optical tracker, \{\textit{H}\} head of the subject, and \{\textit{Hr}\} rigid body mark on the head. \{$R\rightarrow E$\} is obtained by robot sensors, \{$O\rightarrow Cr$\} and \{$O\rightarrow Hr$\} by the optical tracker, \{$E\rightarrow Cr$\}, \{$Cr\rightarrow C$\}, \{$Hr\rightarrow H$\} by registration and calibration (Section \ref{sec:registration}), and \{$H\rightarrow b$\} by pre-operative planning (Section \ref{sec:toolplan}). \textbf{TMS and EMG setup} shows the TMS stimulation controller and data recording devices and their connections. More details are provided in Section \ref{sec:overview}. This figure contains graphical elements licensed from BioRender.com. }
    \label{fig:architecture}
\end{figure*}

\section{Introduction}

TMS is a common non-invasive technique used in neuroscience applications. It can be used in research to map brain function, investigate central disease mechanisms, and study spatial orientation perception \cite{otero2018exploring}. Clinically, TMS is an FDA-approved procedure to treat depression and obsessive-compulsive disorder. TMS is also reported to be a potential treatment for a variety of disorders such as anxiety \cite{balderston2020mechanistic}, Parkinson’s disease \cite{mi2020repetitive}, and Alzheimer’s disease \cite{weiler2020transcranial}. TMS uses brief and rapidly changing magnetic fields, created by a strong current in the TMS coil resting on the head of the subject, to produce an electric field in the brain that depolarizes neurons and modulates their activities \cite{lefaucheur2019transcranial}. A basic TMS equipment \cite{valero2017transcranial} consists of (1) a TMS coil made of loops of copper wires to generate rapidly changing magnetic fields, (2) a control machine to configure the stimulation parameters such as frequency and intensity of the current, and the trigger to release the stimulation pulses, and (3) capacitors to accumulate the charges from the power source. Realtime stereotaxic neuronavigation systems may be used to visualize and correctly place the TMS coil above the target cortical area in the brain of the subject \cite{ruohonen2010navigated}. It is achieved after a process of medical image registration to an optically tracked rigid body marker on the head of the subject. Most TMS studies have been performed using conventional manual methods, where an expert operator aligns the coil manually with the guidance of a neuronavigation system. Recent advancements in robotics have been adopted for medical procedures \cite{dupont2021decade}, and robot-assisted procedures may demonstrate higher accuracy and more stable outcomes than those carried out by expert operators.

There have been some attempts to perform robot-assisted TMS in the literature. These studies have primarily focused on the workflow \cite{lancaster2004evaluation, meincke2016automated, liu2023toward}, robotic system design \cite{yi2012design, ginhoux2013custom, liu2023toward, zhang2024realtime}, and calibration methods \cite{noccaro2021development, liu2024gbec}. However, the exploration of precisely controlled tool poses with respect to a fine-segmented brain model is limited. The approaching pose of the TMS coil may impact the stimulation outcome since a pose difference alters the distribution of the induced electric field on the brain anatomy. Significantly, the recent studies in computational simulation of TMS and TES (Transcranial Electric Stimulation) reveal the impact on the induced electric field by anatomical details \cite{saturnino2019simnibs, mantell2023anatomical}. The impact is more profound in individual anatomy differences to the electric field \cite{li2015contribution}. The above all imply that fine segmentation of the brain images and accurate pose control are needed in rigorous TMS studies. 

A critical reason for this research gap is the absence of a standardized, precisely quantified reporting method for the tool poses using fine segmentation. In the existing neuro-navigation systems (Fig. \ref{fig:preopplan}a and b), the tool poses in the experiments are not reported with respect to the true boundary or a fine segmentation of the brain, and the tool poses of a TMS session are assigned by subjective decisions. The procedure usually involves manually dropping markers on the 3D brain image and tweaking the approaching angles of the TMS coil using built-in control options in the software. Thus, deciding the pose, especially the orientation, in the preoperative planning procedure (dropping and tweaking the markers) is subjective and depends on the operator's experience. In clinical practice, targeting the vicinity of the presumed cortical area is considered sufficient. There is a lack of guidelines to provide a pinpointed coil pose, including both position and orientation in the practical use \cite{rossi2009safety}. Before the adoption of robots in TMS, the accuracy of manual alignment had not sufficiently justified establishing a standard reference (true zero) with respect to the segmented brain image for reporting coil poses. Achieving accurate reporting would be challenging even though a reference was given, simply because of human errors when aligning the TMS coil \cite{sack2009optimizing}. However, with the enhanced actuation accuracy provided by medical robots, reporting with such precision is now justifiable. To this end, our main contribution is to develop a robot-assisted TMS system with the following features: 
\begin{itemize}
    \item The system uses fine brain segmentation for the neuronavigation system in TMS. Heuristics of automatically calculating the TMS coil pose are provided to establish a standardized and precise report method (Fig. \ref{fig:preopplan}).
    \item Robotized coil placement ensures accurate and stable coil pose on the fine brain model populated by the heuristics in the planning stage (Fig. \ref{fig:architecture}), validated by our results from custom-made magnetic field sensors and preliminary human subject study.
\end{itemize}

\begin{figure}[ht]
    \centering
    \includegraphics[width=\linewidth]{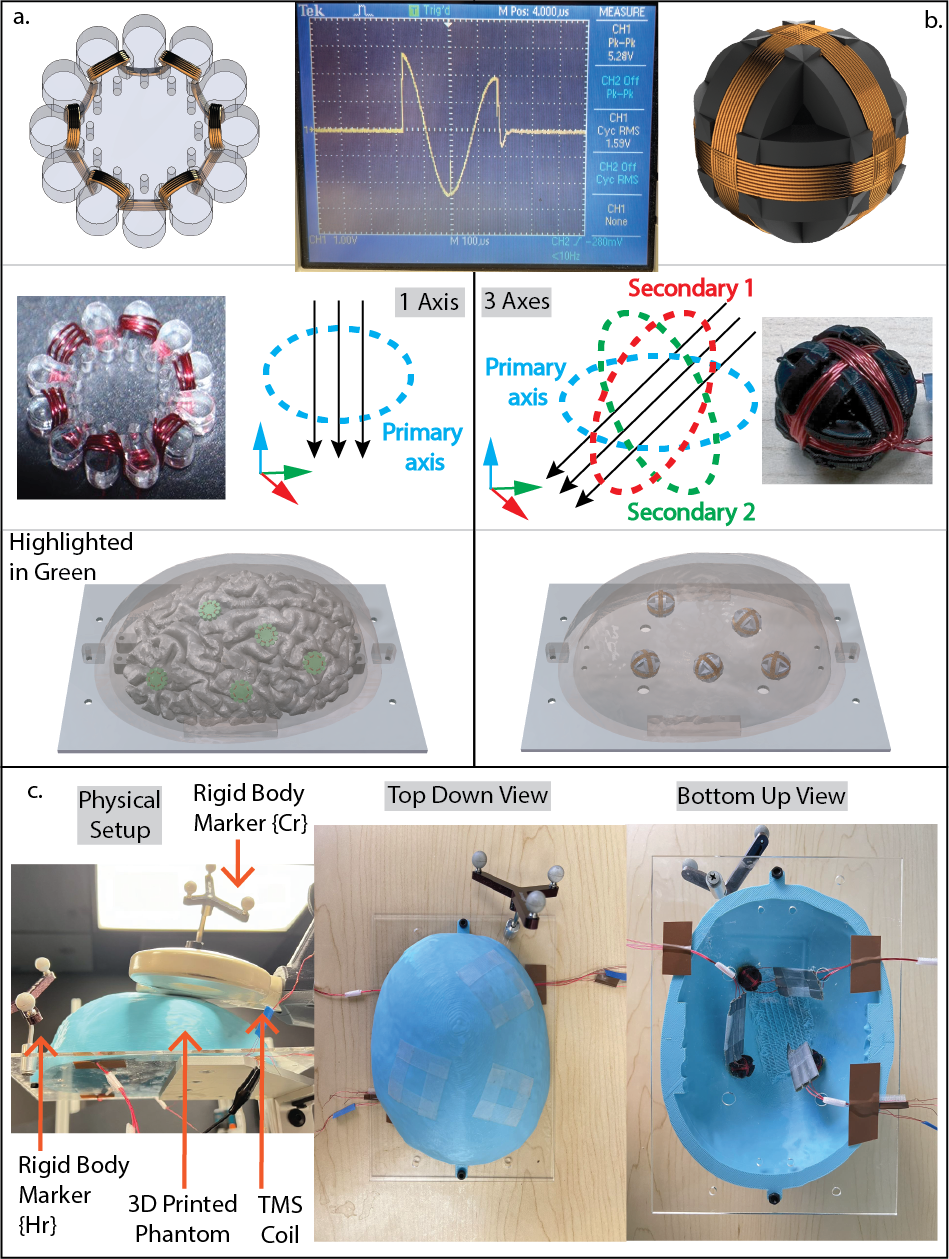}
    \caption{The design and the recording methods of the 2D/3D magnetic field sensors. \textbf{Panel a.} is the design of the 2D magnetic field sensor and a setup using a head phantom. The 2D sensor can measure one axis of the applied magnetic field, and the measurement is a scalar. In the phantom, the brain can be used as a localization tool so that the location of an attached 2D sensor is known when covered by the 3D-printed skin shell. An example of the oscilloscope measurement is shown between panels a. and b. \textbf{Panel b.} is the design of the 3D sensor that can measure all 3 axes of an applied magnetic field in space. The measurement results can be represented in vectors. \textbf{Panel c.} shows a physical setup using the proposed sensors to measure the magnetic field generated by a TMS coil actuated by our robotic system (robotic arm not shown). Rigid body markers are attached for the purpose of navigation. }
    \label{fig:sensors}
\end{figure}

\section{Method}

\subsection{Robot-Assisted System Overview}\label{sec:overview}

The system used here evolves from previous work in \cite{liu2023toward}. The robotic system consists of a user interface on a Linux desktop, a robotic manipulator, a robot controller machine, a TMS controller machine, and a figure-8-shaped TMS coil. The robotic system setup and probing devices are illustrated in Fig. \ref{fig:architecture}. In the user interface, a medical image processing module, a robot command module, and a visualization module are provided, serving as the neuronavigation component in our system. All user inputs in the preoperative planning steps are entered from the \textit{medical image processing module}, where the operator can perform medical image importing, skin and brain selection (Section \ref{sec:segmentation}), medical image-anatomy registration (Section \ref{sec:registration}), registration error estimation, the planning of TMS coil pose (Section \ref{sec:toolplan}), and hotspot hunting. After the planning of the coil's pose with respect to the brain, the operator may send commands through the \textit{robot command module} in the user interface. In this module, the user can control the robot's connection, receive the robot's state, command the robot end-effector with the TMS coil to the planned pose, and manually adjust the current end-effector pose. During a TMS experiment, the \textit{visualization module} allows for adjustments to the display of medical images, including the brain and skin models segmented from the MRI. It also shows the real-time position and orientation of the TMS coil with respect to the head of the subject. Additional details of the implementation of each software module can be found in \cite{liu2023toward}. 

A 7-axis robot arm KUKA LBR iiwa 7 (KUKA, Augsburg, Germany) is used in our experiments but can be replaced given the corresponding software API. The robot has 820 mm of reach and is advertised to have 0.1 mm repeatability. The model is widely used in medical and surgical robotics research. The TMS controller machine is a Magstim (Magstim Co., Whiteland, UK) Rapid$^2$ stimulator. The TMS coil is a Magstim figure-8-shaped and flat-surfaced coil capable of focused magnetic field generation \cite{ueno2021figure}. Additional hardware is used for data collection in our brain phantom test and the preliminary human subject experiments. For the brain phantom test, we design magnetic field sensors and use an oscilloscope, Tektronix TDS2012, to probe the intensity of the induced current (Section \ref{sec:fieldsensor}).

\subsection{Medical Image Segmentation}\label{sec:segmentation}

Existing neuronavigation systems available commercially usually use the curvilinear model of the brain. This model is obtained from individual subjects' MRI to visualize the cortical ribbon \cite{comeau2014} but lacks details of the true boundary of the pial surface of the brain. A comparison between the brain surface obtained from an existing commercial system and ours is in Fig. \ref{fig:preopplan}a and b. The cortex and the skin are reconstructed from the fine segmentation of the brain derived from FreeSurfer \cite{fischl2012freesurfer}.

As the current computational research reveals, microanatomy affects the distribution of the brain's induced electric fields \cite{opitz2011brain, mantell2023anatomical}. It has also been pointed out in separate studies that individual anatomy difference impacts the electric field propagation \cite{li2015contribution} and the variation of stimulation effect \cite{guerra2020solutions}. The TMS simulation studies require a fine segmentation because of the underlying algorithm \cite{saturnino2019simnibs}, where the electric field simulation in materials requires setting the boundary condition. On each side of the boundary, the material property changes. The coefficients, such as tissue conductivity, should be assigned accordingly. In light of these studies, the models used in the neuronavigation system should accurately represent the true geometric shape of the brain since the study of the mechanism of TMS relies on understanding the connection between the induced electric field and the modulated physiological effect.

\subsection{Registration and Calibration}\label{sec:registration}

The robotic arm moves the TMS coil accurately to a planned pose. A complete kinematic chain should be established to calculate the desired robotic arm joint angles. As shown in the Operation Scene panel of Fig. \ref{fig:architecture}, the transformations \{$Cr$ $\rightarrow$ $E$\}, \{$Cr$ $\rightarrow$ $C$\}, and \{$Hr$ $\rightarrow$ $H$\} are not directly obtained using sensors or pre-operative planning. Instead, they need to be obtained by registration or calibration. The processes of getting these transformations are as follows.

The transformation \{$Hr$ $\rightarrow$ $H$\} is to establish the geometrical relation between the medical image and the subject's head so that an anatomical location inside the brain of the subject is known with respect to a reference frame in the physical setup. In this case, the reference frame is $Hr$ on the rigid body marker on the head. The process is called registration. We use pair-point registration \cite{Arun1987} in our experiments. In the process, we use a tracked probe to touch some anatomical landmarks on the head of the subject, usually including the tip of the nose, nasion, mid-point between eyes, and tragi of both ears. The coordinates of these landmarks are easily identifiable in an MRI. We can then obtain two point clouds, one with respect to the medical image $H$, and the other with respect to $Hr$. The transformation between them can be calculated using \cite{Arun1987}. The pair-point registration in human subject experiments may be affected by the deformation of the skin and discrepancies between the reconstructed MRI and the head.  To ensure an accurate result, we use the Iterative Closest Point (ICP) algorithm \cite{besl1992method} to fine-tune the transformation. The algorithm requires the mesh model of the head and a point cloud collected with respect to $Hr$. In our human subject experiments, we reject any registration results with an average pair-point residual higher than 6 mm, and an average ICP residual higher than 2 mm.

We use a coil center calibration method proposed in \cite{liu2022inside} to obtain the transformation \{$Cr$ $\rightarrow$ $C$\}, which has sub-millimeter accuracy \cite{liu2022inside}. Geometry-based hand-eye calibration (GBEC) is used to calibrate the transformation \{$Cr$ $\rightarrow$ $E$\} \cite{liu2024gbec}, also reported to have sub-millimeter accuracy.

\subsection{Pre-Operative Planning}\label{sec:toolplan}

The trajectory of the placement of the TMS coil is determined by pre-operative planning in the navigation system. In the kinematic chain (Fig. \ref{fig:architecture} \textit{Operation Scene} panel), the obtained pose from pre-operative planning corresponds to \{$H\rightarrow b$\}. Here, we introduce the mathematical formulation of pose calculation and TMS coil plan strategies.

\subsubsection{Extracting Pose from a Curved Surface}\label{sec:toolplansurface} 
A pose in space is a 6 degrees of freedom (DOF) quantity, consisting of translation $(x_c, y_c, z_c)$ and orientation $(r_x, r_y, r_z)$. The translation component can be easily defined by a center point in space, $p_c$, and the rotation around the $y$-axis, $r_y$, can be determined by a reference or ``tail direction point'', $p_t$, as illustrated in Fig. \ref{fig:preopplan}c and f. We can formulate the problem of obtaining the remaining degrees of freedom in 3D space as a plane determination problem. Specifically, given a point $p$, we use two additional points $p_1$ and $p_2$ in space to determine a plane. The normal vector $\mathbf{n}$ to this plane, calculated as:
\begin{equation}
\mathbf{n} = \frac{(p_1 - p) \times (p_2 - p)}{| (p_1 - p) \times (p_2 - p) |}    
\end{equation}
Then, we use $\mathbf{n}$ (defines z-axis) and the tail vector $\mathbf{y} = p_t - p$ ($p_t$ can be selected from $p_1$ and $p_2$) to get $\mathbf{x}$ = $\mathbf{y}\times \mathbf{n}$. The rotation matrix of the pose can then be represented by:
\begin{equation}
    R=(\mathbf{x},\mathbf{y},\mathbf{n})
\end{equation}
So, the final pose is $(p_c,R)$. The center point $p_c$ and the points $p,p_1,p_2$, that we used to obtain the rotation matrix above, are the 4 points used in the ``4-Point Constraint'' strategy to extract a pose from dropping points on a curved surface. 

We can also select a point from $p,p_1,p_2$ to serve as $p_c$, which gives us the ``3-Point Constraint''. To further simplify the method, we can use the 3 points from a triangle on the mesh model instead of manually dropping all required points. That way, only one center point and one tail direction point are required, which is the ``2-Point Constraint''. In this strategy, we search for the closest triangle to the center point out of all the triangles in the mesh model. Then, we use the vertices of the triangle to perform the same calculation described above to get the final pose.

\subsubsection{Tool Pose Plan Strategies}\label{sec:toolplanstrategies}

The targeted stimulation is in the subject's brain, but the TMS coil is placed on the scalp. Therefore, the pre-operative planning should project the targeted pose in the brain to a pose on the scalp. To that end, there can be many heuristics to determine how such a projection can be made. Most intuitively, we can manually place points on the skin surface while simultaneously viewing the underlying target cortex by adjusting the transparency of the skin. This is the ``Free Skin Pose'' strategy in Fig. \ref{fig:preopplan}d. Like conventional neuro-navigation systems, the pose calculation does not directly consider the cortex geometry. The geometry information is only indirectly considered in the subjective decision of dropping the target points on the skin. This strategy still relies on subjective decisions, but in this work, we aim to ensure the process of robotic TMS is as deterministic as possible so that standards can be established. 

Thus, we provide two better heuristics, also shown in Fig. \ref{fig:preopplan}d., namely, ``Restricted Cortex Pose'' and ``Closest Skin Pose''. In obtaining a Restricted Cortex Pose, we first extract a target pose from the curved surface of the region of interest on the brain, using any of the strategies described in the previous section. Next, we use the z-axis (perpendicular to the brain surface) of the extracted pose to get the intersection of the skin surface. The intersecting point is used as the center point. The final pose is then defined by this center point on the skin (translation) and the tangential plane on the brain surface (orientation). In practice, since there is only one tangential point on the assumed convex shape of the scalp skin to any plane, it can be difficult to align the TMS coil to a plane that is tangential to the brain shape without a collision of part of the TMS coil to the skin. Just as shown in Fig. \ref{fig:preopplan}d. where the plane is a secant plane cutting through the skin. Thus, we may use the next strategy, which ensures a tangential plane on the skin. In the process of obtaining Closest Skin Pose, again, we first get a target pose on the brain surface. Then, we search for the closest mesh triangle on the skin surface. The final pose is defined by both the center point and the orientation of the mesh triangle. In this strategy, the coil is less aligned with the brain.

It is worth noting that any strategy introduced in this section is fully deterministic because the derivation of the poses is entirely up to the algorithm with minimal user inputs. This means that the process is more repeatable than that of the existing systems, where subjective decisions have to be made when selecting the orientation of the pose. 

\subsection{2D and 3D Magnetic Field Sensor}\label{sec:fieldsensor}

Sending the same pulses to a human subject in the same pose may not induce the same effect due to a long chain of physiological effects, as our understanding of the stimulation and the underlying mechanisms is still incomplete \cite{spampinato2023motor, bestmann2015uses}. Thus, we use magnetic field sensors, a more deterministic method, for the validation of this engineering system. The measurement of the magnetic field can be performed using inductive sensors. They are easy to design with winds of insulated magnetic wires. We design inductive sensors that operate based on the principle of electromagnetic induction. When placed in an alternating magnetic field, an alternating current is generated in the turns of the inductive sensor, which an oscilloscope can measure. The frequency of the induced current is the same as that of the magnetic field, with a phase shift.

To probe the magnetic field, we first design a 2D sensor, shown in Fig. \ref{fig:sensors}a. The frame is laser-cut from a 3mm acrylic board to support the enameled copper wire. Each sensor is circular-shaped, wrapped around a series of evenly spaced cylindrical elements to provide stable wire support without unintended sliding. The radius of the location of the wiring on the frame is 7.5mm. We use a strand of enameled magnet winding wire sized 30 American Wire Gauge for the sensor wire. Each sensor has 10 turns. Fig. \ref{fig:sensors}a. also shows a phantom setup, where a segmented brain and a skin shell are 3D-printed and assembled on a supporting board. The setup size and relative positions are from an accurate MRI model. The supporting board, also laser-cut from a 3mm acrylic board, contains fastener holes to ensure an accurate position. Using the MRI imported into our navigation system, the brain phantom can locate an attached 2D sensor.

The TMS coil-generated magnetic field is distributed throughout 3D space, and at each point, it can be represented by a vector indicating the field's direction and intensity. Thus, we extend the 2D sensor and design a 3D sensor capable of capturing measurements along all three axes. Shown in Fig. \ref{fig:sensors}, the 3D sensor frame is 3D printed and also has a wiring radius of 7.5mm. Three separate wires wrap the sensor in all orthogonal directions. The wires rarely slide because the wrapping of all 3 wires stabilizes each other, so stabilizers like the cylindrical elements in the 2D sensor are unnecessary. In an ideal scenario, the induced current in the tangential direction of the brain surface is stronger. The strongest field is emitted through the center of the TMS coil, and we use the term ``primary direction'' to indicate this direction. In Fig. \ref{fig:sensors} a. and b., the direction is the light blue colored axis. The 2D sensor measures only the primary direction, while the 3D sensor measures the primary direction and two secondary directions.

The 2D sensor and 3D sensor serve different purposes. The 2D sensor is better used in scenarios where only the primary direction is required, and the location of the measurement point is needed before the experiments, as the brain phantom can be used to obtain the location of the attached sensor. For example, the evaluation of a magnetic field simulation where the measurement's location is required. On the contrary, due to the size of the 3D sensor, it is hard to fit it between the skin shell and the brain phantom. In such experiments, only the skin shell is used, and the 3D sensors are glued inside. We cannot obtain the precise location of the 3D sensor based on the MRI. Therefore, the 3D sensor is better used when all 3 axes are required and when the location of the measurement point is not needed before the experiments. For example, in brain mapping experiments where they need to mimic the process of locating an unknown location in the brain.

\section{Experiments and Results}

We test the \textit{accuracy} of the robotic and manual methods by alignment tests on human subjects. In the alignment test, we repeatedly move the TMS coil to a pose generated by our algorithm and log the translational and rotational data. The actuation errors for both methods are the basis of the accuracy analysis. 10 repetitions were performed for each method. The subjects were seated upright by the robot and bit on a fixation bar to avoid excessive head motion. To reach a minimum distance from the coil to the underlying cortex \cite{mcconnell2001transcranial}, we used the Closet Skin Point strategy introduced in Section \ref{sec:toolplanstrategies} and Fig. \ref{fig:preopplan}. The experiments were approved by the Johns Hopkins Institutional Review Board, and written consent was obtained from all subjects. 

For the phantom test, we expect that the better actuation method can induce \textit{more stable peak-to-peak voltage} when holding the coil for a long time. We target a sensor underneath a 3D-printed shell (Fig. \ref{fig:sensors}). Here, we use the 3D sensor because we may observe how the primary and secondary axes react differently. The sensor's location is intentionally unknown, because we aim to have a mock-up TMS session in a head phantom, where optimal location needs to be searched beforehand. For the test, a hotspot-hunting procedure is performed by automatically generating a grid of candidate poses on the phantom head and locating the highest responding pose. Then, the hotspot is used in all following tests. 

Inducing a current in the sensor requires an alternating magnetic field, so we use the repetitive mode in the TMS controller to send trains of pulses instead of single pulses. Each trial is a train of 25 pulses, with the stimulation intensity set to 30 MSO\% (percentage maximum system output) and frequency to 5 Hz. An example of oscilloscope recording is shown in Fig. \ref{fig:sensors}. Next, we compare the robotic and manual actuation to align to the hotspot and hold for 5 minutes while sending 20 trains of pulses. Each train of pulses takes 5 s (= 25 Pulses $\div$ 5 Hz), and the wait time between two consecutive trains is 10s. Two operators participated in the phantom test.

\begin{figure*}[ht]
    \centering
    \includegraphics[width=\textwidth]{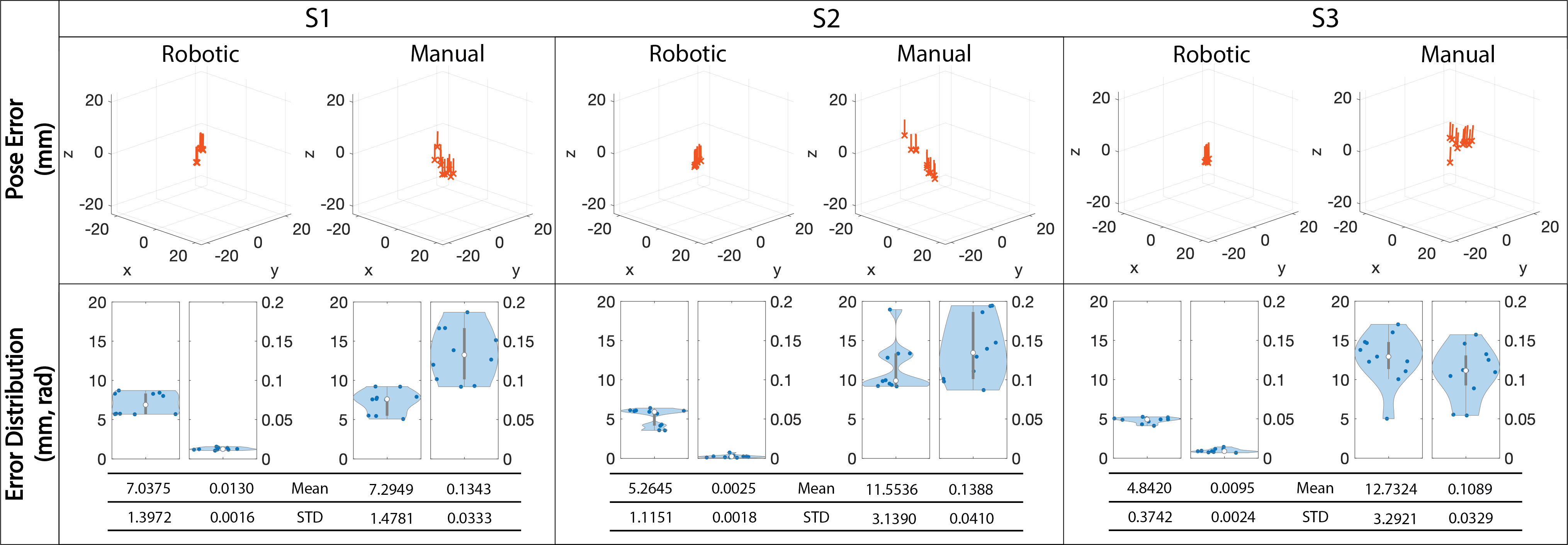}
    \caption{The results from the alignment tests in the human subject study. From left to right, S1-3 are the 3 subjects. The \textbf{Pose Error} panel shows the errors between the planned and measured poses in the 10 alignments for each method. The error transformation is from the planned pose to the measured pose, and all 3 axes are shown. Geometrically, the z-axis is the perpendicular direction of the skin surface, and the y-axis is in the direction of the tail of the TMS coil. The poses are represented in small vectors where the translation is the coordinate of the small cross, and the orientation is the tilt of the vectors. The size of the cluster represents repeatability. The \textbf{Error Distribution} panel shows the violin plots of each group of 10 alignments. The results in this panel are shown in Euclidean distance and the angle component of the angle-axis representation of a 3D rotation. The left sub-column shows the translational errors, and the right shows the rotational errors. The means and the standard deviations are provided in the table at the bottom of each row.}
    \label{fig:PoseOnly}
\end{figure*}

\begin{figure*}[ht]
    \centering
    \includegraphics[width=\linewidth]{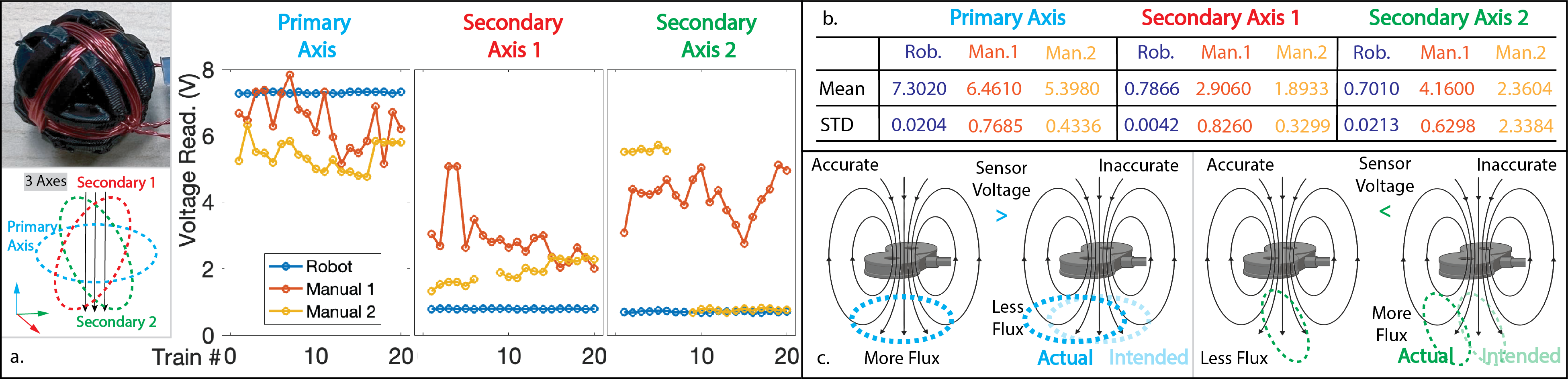}
    \caption{The results of the coil-holding test of the phantom head and 3D sensor voltage measurement. \textbf{Panel a.} shows the sensor voltage measured at the same pose from sending 20 trains of the pulses over 5 minutes. Each sub-panel shows the results in one axis from three actuation methods, robotic, manual operation 1 and 2. The data of manual operator 2 contains missing traces in the secondary axes, and those results are skipped in the plot (gaps in the yellow line).  \textbf{Panel b.} shows the mean and standard deviation, organized by axes and actuation methods. The numbers in the table are color-coded in the same colors as the lines of the different actuation methods in Panel a. \textbf{Panel c.} illustrates that a small sensor displacement causes a change of the received magnetic flux, where the more accurate method to have lower readings in the primary axis, and the less accurate one may have higher readings in the secondary axes. The illustration matches the results in Panel a. This figure contains graphical elements licensed from BioRender.com. }
    \label{fig:SensorVolt}
\end{figure*}

\subsubsection{Robot-Assisted Alignment}
\label{sec:robalign}

Fig. \ref{fig:PoseOnly} shows the pose error and the distribution of the 10 alignments in human subject tests. As a general observation, The results consistently show that the robotic method outperforms the manual method in terms of both translation and rotation errors. In the Pose Error row, the data cluster of the robotic methods in all 3 subjects is smaller than the ones in the manual column, indicating not only that the errors are closer to zero but also that the results are more consistent and repeatable. The values listed in the tables in the Error Distribution support the same observation: The error's means and standard deviations for the robotic method are all smaller than the manual ones. Although this is true for both translation and rotation, it is worth noticing that the average rotational error is significantly lower in the robotic method, with an improvement of 2 orders of magnitude in S2 and an order of magnitude in the other two. More specifically, for S2, the robotic method can achieve an average rotation error of $25 \times 10^{-4}$ rad (0.15$\degree$) while the manual method has an error of $14 \times 10^{-2}$ rad (8.0$\degree$). The results suggest the robot can perform much better in orienting the TMS pose than an experienced human operator.

Due to the small sample size of this preliminary human subject study, it is challenging to draw statistically significant conclusions. Still, the advantage in absolute numbers suggesting the potential benefits of robotic systems in this context is clear.

\subsubsection{Sensor Voltage}

Fig. \ref{fig:SensorVolt}a. shows the readings of the 3D sensor when receiving 20 trains of pulse and holding for 5 minutes. As expected, the voltage \footnote{Please note that the voltage readings from the sensors should be used comparatively to each other, as individual voltage values do not provide sufficient meaningful insights due to their dependence on factors such as the wire winding, the type and frequency of the repetitive pulse, and the pulse's amplitude. } in the robotic method's primary axis is higher than the one using the manual method. This is true for both the manual operators. The mean values in the table in Fig. \ref{fig:SensorVolt}b. confirms this observation. The voltage's standard deviation value of the robotic method is also lower than the manual ones by more than one order of magnitude, indicating the robotic system is more accurate and significantly more stable. This means that, on the primary axis, accurate and stable robotic actuation can translate to a higher induced voltage in the 3D sensor.

On the secondary axes, the standard deviation using the robotic method is one to two orders of magnitude smaller than the manual ones, again, confirming our previous conclusion regarding the stability. In the secondary axes, the results show that the voltages of the robotic method are lower than the manual ones. This is due to the fact that, if the alignment is exact, the readings from the secondary axes should be zero since the magnetic field does not have a horizontal component directly beneath the center of the TMS coil. In contrast, only a vertical component contributes to the primary axes, as labeled by ``Intended'' in Fig. \ref{fig:SensorVolt}c. Thus, a more accurate method should induce a higher voltage in the primary axis but lower in the secondary axes. Although minor, our non-zero results in the robotic method can be explained by a displacement resulting from insufficient hotspot hunting. The direct cause of the voltage drop can be better illustrated in Fig. \ref{fig:SensorVolt}c. Assuming the center of the coil is not directly above the center of the sensor, due to the non-uniformly distributed magnetic field, the magnetic flux passing through the winding on the primary axis is lower than expected, as some of the components of the magnetic field at this displacement are pointing toward the horizontal direction. This, however, will increase the magnetic flux in the secondary axes, which are oriented to receive magnetic flux in the horizontal direction.

\section{Discussion}\label{sec:discussion}

We have taken specific steps to address the engineering factors affecting the stimulation. We have addressed the registration errors in Section \ref{sec:registration} by ensuring the lowest possible registration residual and hand-eye calibration error ($<$ 2mm). The tracking accuracy of our optical system is reported to be submillimeter \cite{koivukangas2013technical}. Our accuracy of coil alignment has been validated in Section \ref{sec:robalign}. By implementing our Closest Skin Point strategy, discussed in Section \ref{sec:toolplanstrategies} and illustrated in Fig. \ref{fig:preopplan}, we have minimized the distance between the coil and the underlying cortex.

Despite our results showing the advantages of the robotic method over the manual method, some limitations remain. First, as discussed in Section \ref{sec:toolplan}, both ``Restricted Cortex Pose'' and ``Closest Skin Pose'' heuristics can be a precise reference in defining the pose of the TMS coil. However, in practice, they can be limited in terms of different aspects. Restricted Cortex Pose does not avoid a collision or pressing of the TMS coil into the head, simply because any point on a convex surface has only one tangential plane. When moved, the tangential plane becomes a second plane slicing through the convex surface. In contrast, the Closest Skin Pose strategy ensures not pressing into the skin by incorporating less brain geometry. The orientation of the automatically calculated pose is tangential to the skin surface, not the brain surface.

The strategy to obtain a reference pose is still an open-ended question. Our three strategies are the most intuitive examples, leveraging the fine segmentation of the brain and skin. More heuristics can be proposed if they minimize randomness and subjective decisions. Possible solutions should also depend on the logic of the needs and determine the available information's trade-off. Although our results are evident, our human subject study is preliminary and has a cohort of 3 subjects. Additional work inviting a larger cohort of subjects and manual operators will investigate individual differences more and can derive more thorough statistical significance.

\section{Conclusion}

In this work, we established a standard for using fine-segmented brain and skin surfaces to populate the TMS coil's target pose automatically. Three heuristics were proposed for different objectives, and precise robotic actuation ensures the execution of the preplanned pose. A human subject study and experiments using customized magnetic sensors were performed to validate our robotic system. Our findings demonstrate that our robotic system reduces positional error by half and improves rotational accuracy by up to two orders of magnitude. The repeatability of these improvements is confirmed through the reduced standard deviation (an order of magnitude) observed across multiple trials. These improvements in actuation accuracy are effectively translated to stimulations, as shown by the higher induced voltage in magnetic field sensors. The stability of the induced voltage over time is also improved by up to two orders of magnitude in terms of standard deviation.

\bibliographystyle{IEEEtran}
\bibliography{bib} 

\end{document}